\documentclass[11pt,a4paper]{article}
\usepackage[hyperref]{acl2020}
\usepackage{times}
\usepackage{latexsym}
\usepackage{latexsym}
\usepackage{amsmath}
\usepackage{graphicx}
\usepackage{url}
\usepackage{color}
\usepackage{enumitem}

\usepackage{microtype}

\aclfinalcopy 

\title{When do Word Embeddings Accurately Reflect Surveys on our Beliefs about People?}

\author{Kenneth Joseph \\
  Computer Science and Engineering \\
  University at Buffalo\\
  Buffalo, NY, 14226 \\
  \texttt{kjoseph@buffalo.edu} \\\And
  Jonathan H. Morgan \\
  University of Applied Sciences \\
  Fachhochschule Potsdam \\
  Potsdam, Germany \\
  \texttt{morgan@fh-potsdam.de} \\}

\date{}

\begin{document}
\maketitle
\begin{abstract}
Social biases are encoded in word embeddings. This presents a unique opportunity to study society historically and at scale, and a unique danger when embeddings are used in downstream applications. Here, we investigate the extent to which publicly-available word embeddings accurately reflect beliefs about certain kinds of people as measured via traditional survey methods. We find that biases found in word embeddings do, on average, closely mirror survey data across seventeen dimensions of social meaning. However, we also find that biases in embeddings are much more reflective of survey data for some dimensions of meaning (e.g. gender) than others (e.g. race), and that we can be highly confident that embedding-based measures reflect survey data only for the most salient biases.  
\end{abstract}

\section{Introduction}

In April of 2015, protests erupted over the murder of Freddie Gray, Jr. Questions about what to call those protesting quickly became the focus of a national debate. In an interview on \emph{CNN} with Erin Burnett,\footnote{http://nymag.com/intelligencer/2015/04/carl-stokes-to-cnn-thug-is-racially-charged.html} Baltimore City Councilman Carl Stokes admonished then-President Barack Obama and then-Mayor Stephanie Rawlings-Blake for using the word \emph{thugs} to refer to the protesters. Burnett challenged Stokes' admonition, claiming the protesters were indeed thugs because ``They know it's wrong to steal and burn.'' Stokes responded by stating the protesters were ``...\emph{children} who have been set aside [and] marginalized.''

\begin{figure}
	\centering
	\includegraphics[width=.48\textwidth]{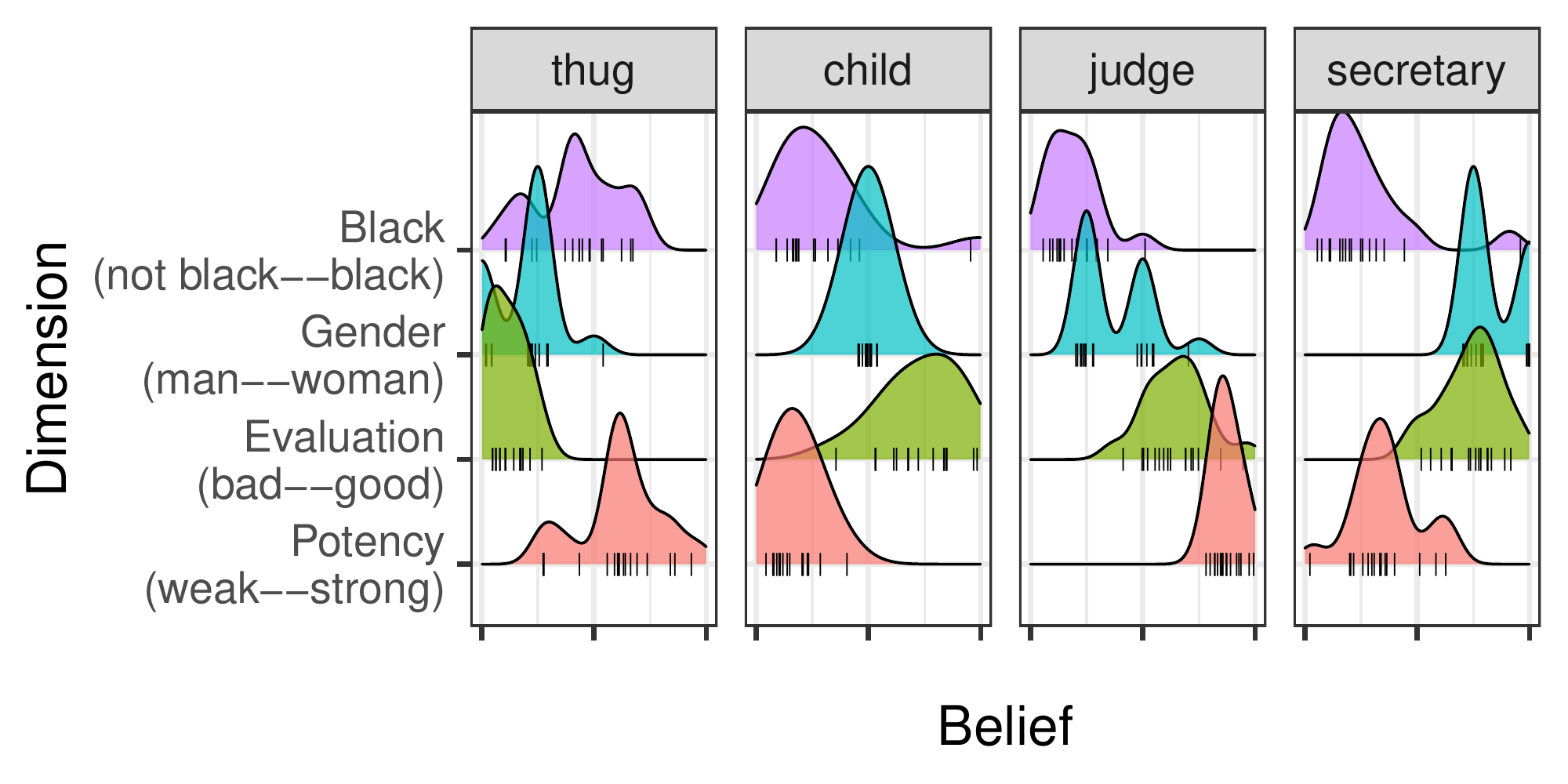}
	\caption{Beliefs (x-axis) about four identities (separate plots) along four dimensions of social meaning (y-axis). Beliefs are displayed as distributions of survey responses. The scale is different for each dimension, e.g. for Evaluation, survey participants responded with their belief as to whether people who held the identity were more likely to be a bad (further left on the x-axis) or good (further right).}
	\label{fig:intro}
\end{figure} 

The argument between Burnett and Stokes is over the way we label people, the meanings of those labels, and the impacts these meanings can have. Councilman Stokes wants to avoid using the label ``thug'' because of its established, negative connotation towards black Americans \cite{dow_deadly_2016}. The survey data collected for this paper, a sample of which is shown in Figure~\ref{fig:intro}, provides further evidence of this association between thugs and black Americans.  Respondents to our survey, on average, expected thugs to be bad, and that approximately 42.4\% of thugs would be black. Of the 57 identities we studied, the only identity perceived to be more black was criminal, at 47.3\%.

The beliefs we have about people who hold particular {\em identities} \cite{mccall_identities_1978} are important, because they often determine the behaviors we take towards people who are labeled with those identities \cite{ridgeway_gender_1999}.\footnote{Different kinds of beliefs about identities have different names. For example, contextualized beliefs are called \emph{impressions} \cite{heise_affect_1987}, and aggregations of beliefs across multiple dimensions of meaning are called \emph{stereotypes} \cite{fiske_model_2002}. The beliefs we study here are typically called \emph{sentiments} or \emph{associations}. However, given the distinct meaning of these terms in NLP, we  use the general term ``belief'' in this paper. This aligns roughly with the generic use of the terms ``bias'' and ``stereotype'' in related NLP literature.} For example, as Councilman Stokes knows, we do not behave the same way towards children as we do towards thugs. This is because, as reflected in Figure~\ref{fig:intro}, people generally believe that children are weak and good, whereas thugs are bad and powerful. This leads us to want to do things like \emph{help} children, versus wanting to \emph{attack} thugs \cite{heise_expressive_2007}. 

However, measuring beliefs is difficult. Traditionally, we have relied on surveys to collect these measurements. But there are tens of thousands of identities \cite{joseph_exploring_2016,heise_self_2010}, and beliefs about them can form along many different \emph{dimensions} of sociocultural meaning (e.g. gender, race, and others displayed in Figure~\ref{fig:intro}).  Measuring beliefs about many identities, on many dimensions, using traditional surveys can therefore be difficult. Further, measuring the evolution of beliefs is often impossible with surveys, because survey data is extremely sparse historically \cite{garg_word_2018}. Finally, measuring how these beliefs change with additional contextual information (e.g. beliefs about specific teachers, rather than teachers in general) is notoriously difficult with survey data \cite{heise_expressive_2007}.

Recognizing these difficulties, scholars have begun to develop NLP tools to measure beliefs about identities historically, at scale, and in context   \cite{joseph_girls_2017,alex2019unsupervised,fast_shirtless_2016-1,garg_word_2018,field2019contextual}. 
Most recent methods derive these measures by manipulating word embeddings.  Studying beliefs enmeshed in word embeddings is also critical because embeddings are widely used in downstream NLP models, which are themselves beginning to label people, for example, as job-worthy or not \cite{de-arteaga_bias_2019}.
Measuring beliefs about people using embeddings therefore serves the dual purpose of understanding human biases and of ensuring such biases are not propelled further along by algorithms.

However, work remains to understand when embedding-based measures of beliefs about identities accurately reflect more traditional survey measures, and why some beliefs may be reflected more accurately than others. The present work combines new and existing survey data with an extensive set of embedding-based measurement strategies to explore this at both the dimension level and the belief level.  At the dimension level, for example, we ask, how well do embeddings capture beliefs about gender, relative to race? And if differences exist, why? Such issues have arisen in existing work, for example, where \citet{garg_word_2018} see correlations of .65 between embedding-based and survey-based measures of beliefs about gender, but only .15 for ethnicity-based beliefs. At the beliefs-level, we ask, for example, how much more accurately do we capture beliefs about the Potency (strength) of thugs, relative to beliefs about the Potency of children? Accuracy at this level is critical for linking historical trends in social behavior to societal-level beliefs about particular identities.

Our primary contributions are as follows:
\begin{itemize}[nolistsep]
	\item We show that \emph{what} we measure is more important than 
	\emph{how} we measure it in determining the correlation between embedding-based and survey-based measures of beliefs about people.
	\item At the dimension level, the beliefs we measure most accurately are also the \emph{most important for how we label others}.
	\item At the belief level, assuming we can identify a good measurement model, embedding-based measures are significantly more accurate for more extreme, and more agreed upon, beliefs.
\end{itemize}
All code and data necessary to replicate the analysis in this article can be found at \url{https://github.com/kennyjoseph/embedding_impressions}.

\section{Related Work}\label{sec:lit_rev}

Our work is grounded in literature on measuring beliefs about identities in social psychology in general and, more specifically, via word embeddings.  We address these two literatures separately here.

\subsection{Belief Measurement in Social Psychology}\label{sec:soc_lit}
 
A common approach for measuring beliefs about specific identities is to assume a \emph{dimensional representation}---that is, to assume a set of distinct dimensions of social meaning can be used to characterize how we think and feel about someone that holds a particular identity.  From this dimensional perspective, two primary questions arise.

First, what are the dimensions along which beliefs form? Social psychologists have identified three classes of important dimensions: traits, affective meanings, and semantic associations.  Traits represent visible---although also socioculturally defined---characteristics like age, gender, and race 
\cite{freeman_dynamic_2011}.  Affective dimensions of social meaning represent how we feel about a given person and/or identity \cite{todorov_social_2015,fiske_model_2002,heise_expressive_2007}. Here, we use the three affective dimensions proposed by \citet{heise_expressive_2007} and that are popular in sociology \cite{rogers_affective_2013}--- Evaluation (goodness/badness), Potency (strength/weakness), and Activity (active/passive).  Finally, social psychologists often characterize beliefs about  identities in terms of semantic associations to particular concepts \cite{freeman_dynamic_2011} or institutions \cite{heise_self_2010}. For example, people link the identities brother and sister together because they are both associated with the family institution.  
In the present work, we collect beliefs for seventeen different dimensions of social meaning, incorporating age, race, gender, evaluation, potency, activity, and six institutional associations.

Second, given a theorized dimension of meaning, how should we measure society-wide beliefs about where particular identities lie on that dimension?  
Here, we adopt perhaps the most common approach, which uses  \emph{semantic differential} scales on surveys \cite{osgood_cross-cultural_1975}. The semantic differential technique asks respondents to place an identity on a sliding scale with two opposing concepts (e.g. weak and strong, see the example in Figure~\ref{fig:survey_intro}A).   Finally, it is worth noting that here, like in most social psychology research, we assume that responses from survey participants generalize to American culture writ large. This assumption is built on the well-established culture-as-consensus paradigm in psychological anthropology \cite{karabatsos_markov_2003,batchelder_test_1988}, and empirical work showing that people tend to agree on the vast majority of their beliefs about people \cite{heise_expressive_2007}.  Nonetheless, many counter-examples exist \cite{berger_status_1992,smith-lovin_affect_1992}. We leave questions about how to address these issues to future work.

\subsection{Measuring beliefs with embeddings}

Embedding-based approaches to measuring beliefs typically follow a three step process of corpus/embedding selection, dimension selection, and word position measurement.  

{\bf Corpus/Embedding Selection} Several recent works have argued that the corpus used can impact measures of beliefs about people derived from word embeddings \cite{lauscher_are_2019,mirzaev_considerations_2019,sweeney_transparent_2019}. For example, \citet{brunet_understanding_2019} show how to reduce gender bias in embeddings by removing particular documents from a corpus. However, several others have shown that in their analyses, the corpus used does \emph{not} significantly impact results \cite{spirling2019word,garg_word_2018,kozlowski_geometry_2018,caliskan_semantics_2017}. Differences in the embedding model used have also been observed to impact measurements \cite{chaloner_measuring_2019}. Again, though, robustness checks from other studies suggest a limited effect beyond the somewhat general hyperparameters of window size and the number of dimensions estimated \cite{garg_word_2018,kozlowski_geometry_2018}. 

{\bf Dimension Selection} To measure beliefs, one first must select a dimension along which the belief is assumed to be held. Much of the literature has focused on dimensions related to gender or race. Others, however, have seen value in moving beyond these dimensions \cite{agarwal-etal-2019-word,sweeney_transparent_2019}. Most relevant is the work of \citet{kozlowski_geometry_2018}, who study the association of 59 concepts across 20 different dimensions of sociocultural meaning, and that of \citet{An2018SemAxisAL}, who induce 732 different dimensions using WordNet to study contextual effects of linguistic meaning. While neither work focuses heavily on identities, these efforts compliment our goal of studying a broad range of dimensions of social meaning.

Scholars then identify a \emph{direction} within the embedding that represents this dimension.  To do so, an approach similar to the semantic differential idea is used. Terms are selected to represent the two ends of the dimension. For example, to identify the gender direction, words at one end might be ``he'' and ``him'', and words at the other end, ``she'' and ``her''. 
Scholarship varies on how these \emph{dimension-inducing word sets} are selected.  For example, several scholars have used demographically gendered and/or racialized names \cite{bolukbasi_man_2016-2,caliskan_semantics_2017}, while others have relied on careful extraction of concepts from dictionaries and thesauri \cite{kozlowski_geometry_2018}.  \citet{kozlowski_geometry_2018} find that having more words at each end generally provides better measurements, and others have found a need to use frequently occurring terms \cite{ethayarajh-etal-2019-understanding,brunet_understanding_2019}. 
Beyond these observations, however, scholars have generally found stable results as long as reasonable word sets are selected.

{\bf Word Position Measurement} Finally, the position of each identity along this direction must be identified.  Doing so entails two major decisions. First, how should one quantify the direction, given the dimension-inducing words?  For example, \citet{bolukbasi_man_2016-2} identify the direction by taking the first dimension of a PCA on the full set of direction words.  
Second, how should one define the position of points along this line? For example, several works use the cosine similarity between the identified ``bias direction'' and the embedding of each identity.  Scholars have also recently proposed supervised methods for word position measurement \cite{sweeney_transparent_2019,agarwal-etal-2019-word}. Such approaches are important, but assume the existence of some training data, which may or may not be available in certain measurement contexts. We therefore do not explore these methods further in the present work.

In sum, using embeddings to measure beliefs requires a series of decisions, the impacts of which are still debated. Below, we provide the most comprehensive study to date on the importance of these decisions on measurement quality.

\section{Survey Data}

We collect two new survey datasets for this paper. The first measures beliefs about the 57 identities\footnote{Because not all embedding models account for bigrams, we removed ``police officer'' from all analyses in this paper. However, for future purposes, we include it in our description of the data here.} in Table~\ref{tab:id_s2} on seventeen dimensions of social meaning described below. The second measures the ways in which a set of survey respondents label people with these identities in hypothetical social situations.  

We used a cluster-based approach to select the 57 identities we study. We study nine sets of six identities, where each set has been shown in prior work to be related in some way. Five of the sets are characterized by a salient association to a specific institution described by \citet{heise_self_2010}. Three sets are characterized by salient trait similarities and differences on gender, age or race/ethnicity. And one set of identities is included where all identities have strong negative Evaluations. Finally, we include three random identities as a mechanism for comparison in other work not described here.  For further details on the selection criteria, survey populations, and results, see the Appendix.

\begin{figure}[t]
	\begin{tabular}{c}
	\includegraphics[width=.48\textwidth]{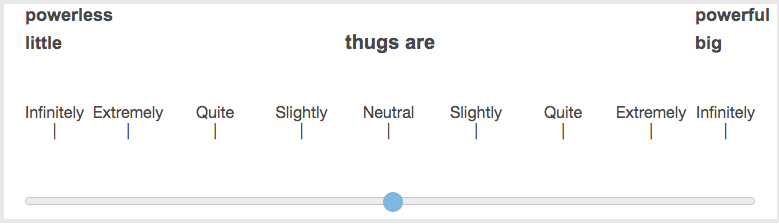} \\
	(a) \\
	\includegraphics[width=.48\textwidth]{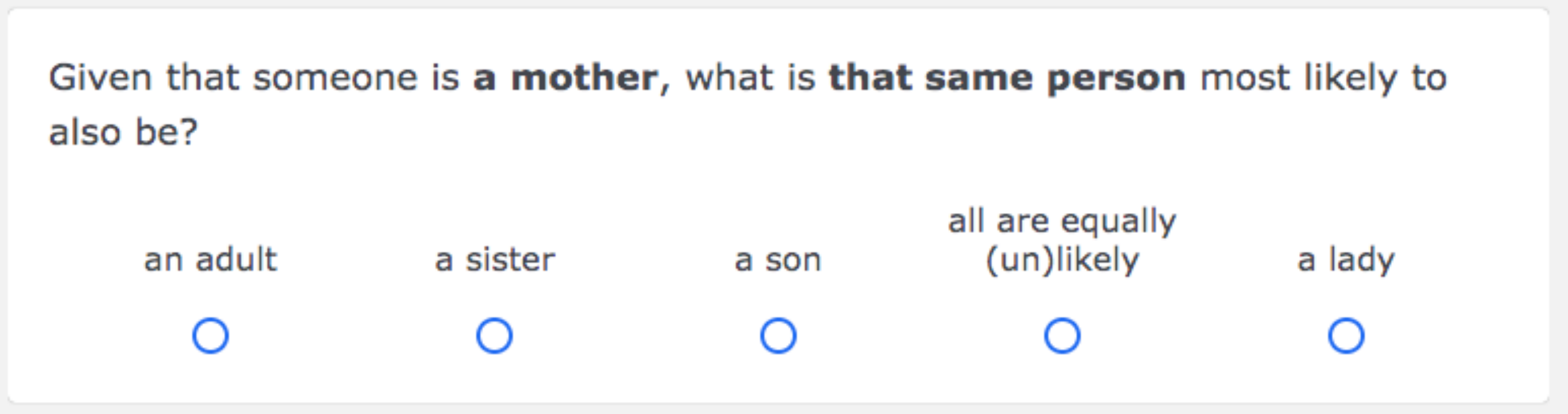} \\
	(b) \\
	\end{tabular}
	\caption{A) Example of a semantic differential question used to measure beliefs about identities (here, for the identity ``thug'' on the Evaluation dimension); B) Example of a hypothetical identity labeling question used to evaluate the importance of different dimensions.}
	\label{fig:survey_intro}
\end{figure}

 \begin{table}[t]
  \small
    \centering
	\begin{tabular}{p{1.4cm}p{5.2cm}}
{\bf Dimension} & {\bf Identities}  \\ \hline
Politics & conservative, Democrat, liberal, Republican, politician, senator \\ \hline
Family & brother, sister, daughter, son, father, mother \\ \hline
Law &  judge, criminal, lawyer, witness, cop, police officer \\\hline
Medicine & doctor, physician, surgeon, nurse, patient, dentist \\ \hline
Business & executive, consultant, secretary, intern, banker, boss \\ \hline
Gender & woman, guy, girl, boy, man, lady \\ \hline
Age & teenager, kid, child, toddler, adult, minor \\  \hline
Race \& Ethnicity & black, white, Hispanic, Asian, Arab, American\\  \hline
Negative Evaluation & thug, idiot, jerk, goon, punk, bully\\ \hline
Random &  principal, scientist, coach \\ \hline
	\end{tabular}
	\caption{The 57 identities we collect data on.  Note that the dimensions used for sampling do not include all dimensions used in our belief measurement study.}
	\label{tab:id_s2}
\end{table}

\subsection{New Belief Measurement Data}

 We collected survey data on beliefs about identities from 247 respondents on Amazon's Mechanical Turk.  Each survey respondent provided responses for four different, randomly selected identities. Each identity was given to a total of 15 different respondents.  For each identity, we asked a set of seven questions, some of which had multiple subparts.  Following prior work, beliefs for affective dimensions were solicited using a slider-based Likert scale.  For the Evaluative dimension, the opposing ends of the Likert scale were labeled ``bad, awful'' and ``good, nice''. For the Potency dimension, ``powerless, little'' and ``powerful, big'', and for the Activity dimension, ``slow, quiet, inactive'' and ``fast, noisy, active''. See \citet{heise_surveying_2010} for more details on the development of these questions. The fourth and fifth question used Likert scales to measure beliefs about age and gender, with ends representing ``young'' and ``old'' and ``Always male'' and ``Always female,'' respectively.
 
The sixth question asked ``Of all [some identity, e.g., bullies], what percentage of them do you think are...'' and then provided one slider each for the following ethnic/racial categories drawn from the planned 2020 Census: White, Hispanic or Latino, Asian, Middle Eastern, and Black.  The seventh question, modeled after the association-based measures from \citet{hill_simlex-999_2014}, asked ``To what extent does thinking about [some identity, e.g., bullies] lead you to think about...'' and then provided a slider for the following institutional settings: family, politics, (criminal) justice, medicine, business, education, and religion.  Each slider had qualitative labels ranging from ``Not at all'', to ``Somewhat'', to ``Immediate response''.  
 
\subsection{Identity Labeling Data}

We collect responses from 402 participants to a pair of identity labeling tasks.\footnote{These identity labeling questions are similar to, but distinct from, those used in our prior work focused on the impact of semantic associations and semantic similarity on identity labeling decisions \cite{joseph_relating_2016}.} Note that these respondents are \emph{different than those who provided the belief measurements}. Each participant answered a set of 40 hypothetical identity labeling questions. Questions could be either an \emph{IsA} or a \emph{SeenWith} question.  An example of an IsA question is given in Figure~\ref{fig:intro}B).  SeenWith questions were formatted in the same way, except the question text instead says ``Who would you say is most likely to be seen with a [mother]?'' 

Questions varied on both the identity provided in the text and the identities serving as potential answers. From the 57 identities we study, we create survey questions roughly\footnote{Due to a bug in Qualtrics, a small percentage of questions were not asked or asked more than once. See Appendix for details} as follows: for a given identity, we generate 14 random sets of the 56 other identities; each set contains four identities. We then generate one IsA and one SeenWith question for each of these sets, where these four identities constitute the possible answers to the question, and the given identity is used in the question text. This process is then repeated ten times for each identity. This process generates ten questions for each of the 3,192 identity pairs for each type of question.

\subsection{Belief Measures From Prior Work}

To further substantiate our claims, we ensure our main results hold using three other datasets on beliefs about identities: beliefs about gender for 287 occupational identities from \citet{bolukbasi_man_2016-2}, beliefs about 195 national and occupational identities on the Big Five Personality Traits from \citet{agarwal-etal-2019-word}, and beliefs about 654 identities   on the Evaluation, Potency, and Activity dimensions by \citet{smith-lovin_interpreting_2015}.

\section{Methods}

\begin{table}[t]
\def\arraystretch{1.2}
  \small
    \centering
	\begin{tabular}{p{1.4cm}p{5.6cm}}
{\bf Variable} & {\bf Description} \\ 
$i$ & A social identity (e.g. doctor, author) \\ \hline
$d$ & A dimension of social meaning (e.g. gender) \\ \hline
$r$ & A survey respondent \\ \hline
$S_{d,i,r}$ & A matrix of survey responses to semantic differential measures on a given dimension $d$ for identity $i$ by respondent $r$. Each observation is in $[0,1]$, where 0 and 1 imply something unique for each dimension depending on the ends of the semantic differential scale.\\  \hline
$\overline{S_{d,i,*}}$  & The average belief of identity $i$ on dimension $d$ in the survey data across all respondents \\ \hline
$E$ & A matrix of word embeddings generated from a particular combination of corpus and embedding algorithm. We refer to this as an \emph{embedding model} \\ \hline
$dw$ & A dimension-inducing word set; that is, a set of words that define the ends of a particular dimension of meaning \\ \hline
$wp$ & A word position measurement model, that is, how it is determined where a given identities lies on a given dimension of social meaning.  \\ \hline
$m_{E,dw,wp}(i)$ & An \emph{embedding-based measurement model}. Defined by an embedding model $E$, a dimension-inducing word set $dw$, and a word position measurement model $wp$. Returns a position for $i$ along the induced dimension \\ \hline
\end{tabular}	
\caption{Notation used in outlining our approach.}
\label{tab:notation}
\end{table}

Our primary research question is, how accurately can we recover beliefs measured using surveys with word-embedding based measures?   We study this first at the \emph{dimension level}, i.e., how accurately do embedding-based measures reflect survey data across \emph{a set of} identities on a given dimension of social meaning? We then study accuracy at the \emph{belief level}, i.e., how accurately do embedding-based measures reflect survey data for \emph{specific} identities on specific dimensions? Our approach is straightforward, but is best explained by introducing some additional notation, provided in Table~\ref{tab:notation}.
 
\subsection{Dimension-level analysis}

At the dimension level, we consider first how different factors relating to the measurement itself impact accuracy. We then study why measurements are more accurate for some dimensions than others. We do so by connecting the degree of accuracy for a given dimension to how important that dimension is in how survey respondents select identities for others in our identity labeling task.

\subsubsection{Impact of measurement strategy}

As discussed above, the accuracy of embedding-based measurements may  vary across properties of the dimension being measured, as well as the way in which the embedding-based measurement is constructed. We first study the relative effects of a) the dimension ($d$), b) the embedding model ($E$), c) the dimension-inducing wordset ($dw$), and d) the word position measurement model ($wp$) on the accuracy of embedding-based measurements. As is standard in the literature, we use the Pearson correlation between the mean survey response and the output of the embedding-based measure as our definition of accuracy. That is, for a given dimension $d$, survey dataset $S$, embedding-based measure $m_{E,dw,wp}$, and set of identities of size $I$, we compute the accuracy of the embedding-based measure as the Pearson correlation between $\{\overline{S_{d,i_0,*}},\overline{S_{d,i_1,*}},...,\overline{S_{d,i_I,*}}\}$ and $\{m_{E,dw,wp}(i_0), m_{E,dw,wp}(i_0), ..., m_{E,dw,wp}(i_I)\}$.  We then run a linear regression to understand how accuracy varies across the factors considered.

Our analysis involves all dimensions of social meaning studied in the four survey datasets described above. For embedding models, $E$, we consider twelve different publicly available corpus/embedding combinations from prior work. To construct dimension-inducing wordsets, $dw$, we using one of three approaches. The first is to use the same terms as were placed on the semantic differential scale on the survey (e.g. powerless, powerful, little, big for Potency, as in Figure~\ref{fig:survey_intro}a). In certain cases, we also include a \emph{survey-augmented} condition that extends this wordset using a thesaurus, after discussion amongst authors. Third, where applicable, we use direction-inducing wordsets from \emph{prior work}. Finally, we consider several of the major established approaches in the literature for word position measurement $wp$. We use the approaches from \citet{kozlowski_geometry_2018}, \citet{swinger_what_2019}, \citet{ethayarajh-etal-2019-understanding}, \citet{bolukbasi_man_2016-2}, and \citet{garg_word_2018}.  In the Appendix, we provide full details on the different values of $E$, $dw$, and $wp$ that we consider.

\subsubsection{Explaining variation across dimensions}

As we will show, controlling for $E$, $dw$ and $wp$, there are large differences in accuracy across dimensions.  To better understand these differences across dimension, we compute two measurements. First, \citet{kozlowski_geometry_2018} show that the variance of the survey data on a dimension, that is, $\mathrm{Var}(\overline{S_{d,i_0,*}},\overline{S_{d,i_1,*}},...,\overline{S_{d,i_n,*}})$, is strongly correlated with the accuracy of embedding-based measures. However, they also note that ``high explained variance... reveals little about how these valences are deployed in social life'' (pg. 930). Here, we therefore compute a second measure that connects variance of the survey data on a given dimension to a significant social outcome, how strongly people rely on that dimension when labeling other people. 

To do so, we first construct a $57x17$ matrix $X$ of scaled-and-centered mean survey responses for each identity on each dimension in our survey data, i.e. $X_{i_0,d_0} = \overline{S_{d_0,i_0,*}}$. We then construct an observation with a binary outcome that pairs the identity in the question with each possible answer. The outcome is 1 if the answer was selected, and 0 otherwise. For example, in Figure~\ref{fig:survey_intro}B), the pairings created would be ``mother, adult'', ``mother, sister'', ``mother, son'', and ``mother, lady''. If the respondent answered ``lady'', then the outcomes would be 0, 0, 0, and 1, respectively. The 40.3\% of questions where respondents answered ``all are equally unlikely'' were ignored. In total, we obtained 9,597 responses where the respondent did not answer ``All are equally (un)likely,'' split roughly evenly between SeenWith and IsA questions.

We then train a logistic regression model for IsA and SeenWith questions separately, each with seventeen parameters. For a given observation, the parameters represent the absolute difference between each dimension, e.g. the first parameter is $|X_{i_q,d_0}-X_{i_a,d_0}|$, where $i_q$ is ``mother`` in Figure~\ref{fig:survey_intro}B), $i_a$ is, e.g., ``adult'', and $d_0$ is, e.g., gender. 

In the Appendix, we provide full results for these regressions. Intuitively, larger negative coefficients for a given dimension indicate that the further away two identities are on that dimension, the less likely the respondent is to select them as a pair. For example, we find that Evaluation has a strong negative correlation for IsA questions, indicating that respondents typically do not expect two identities to be assigned to the same person if one identity is perceived to be for ``good people'' and the other for ``bad people''. Positive coefficients imply assortativity on the dimension. For example, for SeenWith questions, Potency has a positive coefficient, implying that we expect powerful identities to be seen with less powerful counterparts. The magnitude of these coefficients represent the importance given to that dimension by survey respondents. We use the maximum of the two coefficients across SeenWith and IsA questions as a measure of this importance. 

\subsection{Belief-level analysis}

We are also interested in accuracy for specific beliefs. For example, how accurately do embedding-based measures reflect survey data on beliefs about the typical age of a boy? As an outcome for this belief-level analysis, we use a ranking task similar to prior work \cite{spirling2019word,kozlowski_geometry_2018}. We describe this outcome by continuing with the example of beliefs about the age of boys. We first compute the set of identities $N$, for which $\overline{S_{age,boy,*}}-se(S_{age,boy,*}) \, \textgreater \, \overline{S_{age,i,*}}+se(S_{age,i,*})$, where $se$ is the standard error function.  That is, $N$ represents all identities we are reasonably confident respondents believed to be younger than $boy$s. We then determine the subset of $N$, $N_c$, where $boy$ is also ranked above those identities in the embedding measure.  We do the same for identities survey respondents said were older than $boy$s, adding these to $N$, and to $N_c$ if they are correctly ranked in the embedding measure. Finally, we use $\frac{N_c}{N}$ to study accuracy at the belief level.

We are interested both in overall levels of accuracy for belief-level measurements, as well as the factors that explain variation in accuracy. We consider four factors that might explain this variation (continuing with the age/boy example): $sd(S_{age,boy,*})$, the distance of $\overline{S_{age,boy,*}}$ to the median over all identities on that dimension, the logged frequency of the identity in a large corpora,\footnote{according to \cite{robyn_speer_2018_1443582}} and the number of synsets for the identity in WordNet. To study the impact of these different factors, we use a generalized additive model with a binomial link function where $\frac{N_c}{N}$ is the outcome and points are weighted by $N$. Finally, as opposed to considering results across all possible $E$, $dw$, and $wp$, we first select those settings that maximize the Pearson correlation for each dimension.

\section{Results}

\subsection{ Dimension-level results}

\begin{figure}[t]
	\centering
	\includegraphics[width=.48\textwidth]{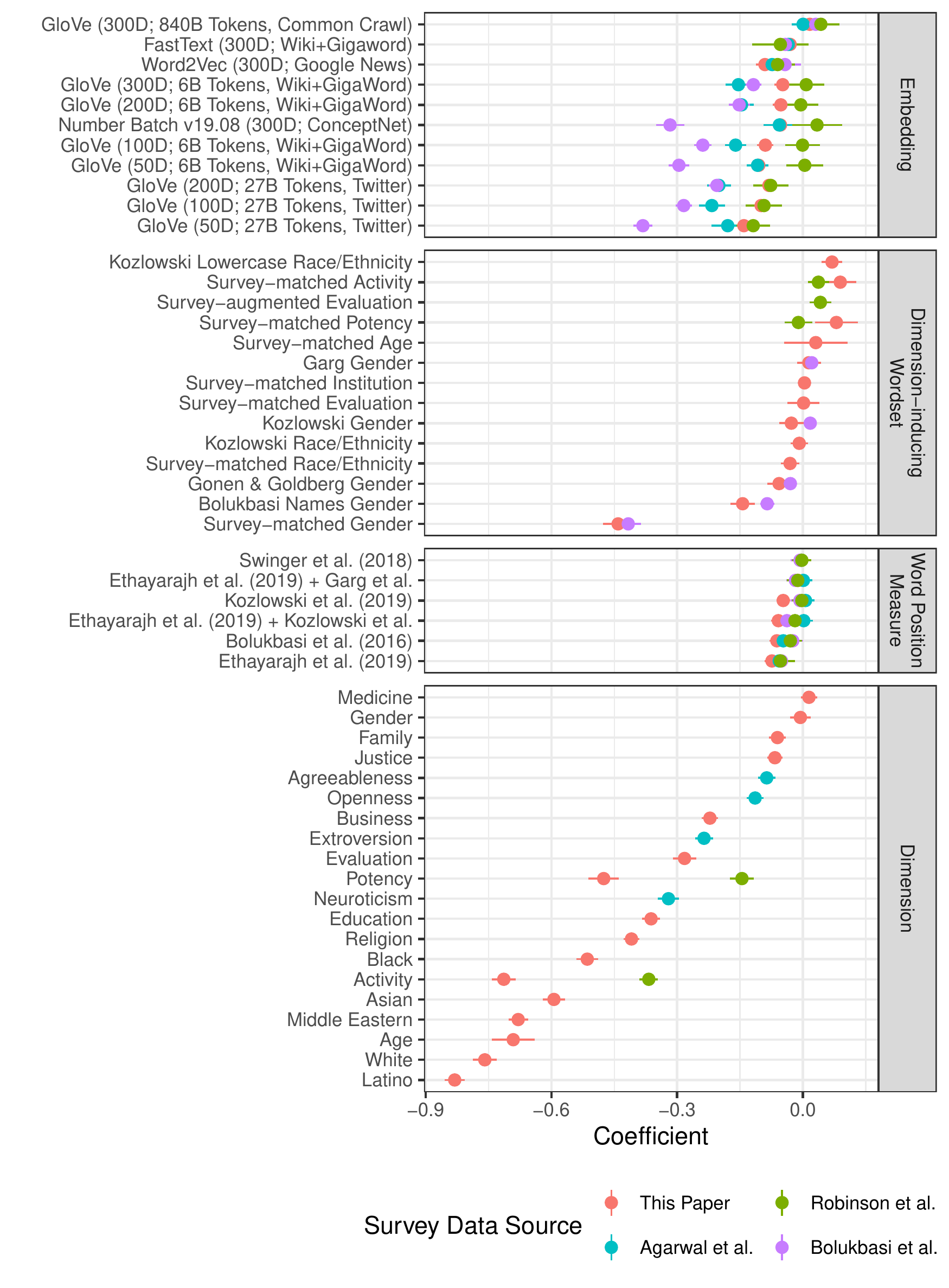}
	\caption{Regression results for the dimension-level analysis. Coefficients for each factor are relevant to a baseline. For the embedding models, the baseline is the FastText 300 dimensional model trained on the Common Crawl. For the dimension-inducing wordset, it is the terms used to define gender by \citet{bolukbasi_man_2016-2}. For word position measurement, it is the approach from \citet{garg_word_2018}, and for dimension, association with Politics.}
	\label{fig:rq1}
\end{figure}

Across all conditions and survey datasets, the Pearson correlation between the embedding and survey measures is 0.32 [.31,.33].  However, considerable variation exists.  Figure~\ref{fig:rq1} presents results of a regression that attempts to explain the sources of this variance (x-axis) and the effects of each source (y-axis). Separate colors represent results from the four different survey datasets analyzed. In general, results are largely consistent across the different datasets, and thus we will not emphasize differences across datasets below.

Figure~\ref{fig:rq1} shows that the embedding model used can decrease correlation by as much as .35. As others have found, this effect decreases when one focuses only on 300-dimensional embeddings. It is worth noting, however, that no embedding model is universally best. For example, nine of the twelve embedding models studied are responsible for producing the highest observed correlation for at least one dimension.

Selection of the dimension-inducing words, $dw$, also has a limited effect. The one exception is when survey-matched words are used for the Gender dimension, where correlations drop by, on average, around 0.5 relative to the ``he/she'' baseline. The fact that using the same words as the semantic differential scale is a terrible choice, but for only one of the seventeen dimensions studied, reflects the fact that selection of $dw$, like elements of other forms of quantitative social science, remains a mix of art and science \cite{sterling2009art}. 

In contrast, even the most scientifically appealing approaches to word position measurement \cite{ethayarajh-etal-2019-understanding} provide marginal gains.  The only consistent observation we draw is that approaches that normalize measurements across dimensions related to the same overarching concept (e.g. that normalize racialized beliefs across all perceived dimensions of race) perform slightly better.  Results thus reflect that the details of measurement are less important than what is being measured. 

Reflecting this same fact, the strongest impacts on correlation between the survey and embedding-based measures come from which dimension is being studied.  
Some of these results reflect the salience of these dimensions in social life.  Associations to institutions, which are most accurately measured on average, are a primary tool we use to sort people into groups \cite{heise_self_2010}. And stronger  correlations between the embedding and survey-based measures for Evaluation and Potency, relative to Activity, reflects the increased importance in affective perceptions of these two dimensions \cite{rogers_affective_2013}.  However, scholars largely agree that trait-based beliefs on gender and race serve as ``default characteristics'' \cite{ridgeway_gender_1999} along which we almost automatically categorize others \cite{todorov_social_2015}. Given their shared salience, why is gender the only trait that can be accurately measured?

\begin{figure}
	\centering
	\includegraphics[width=.48\textwidth]{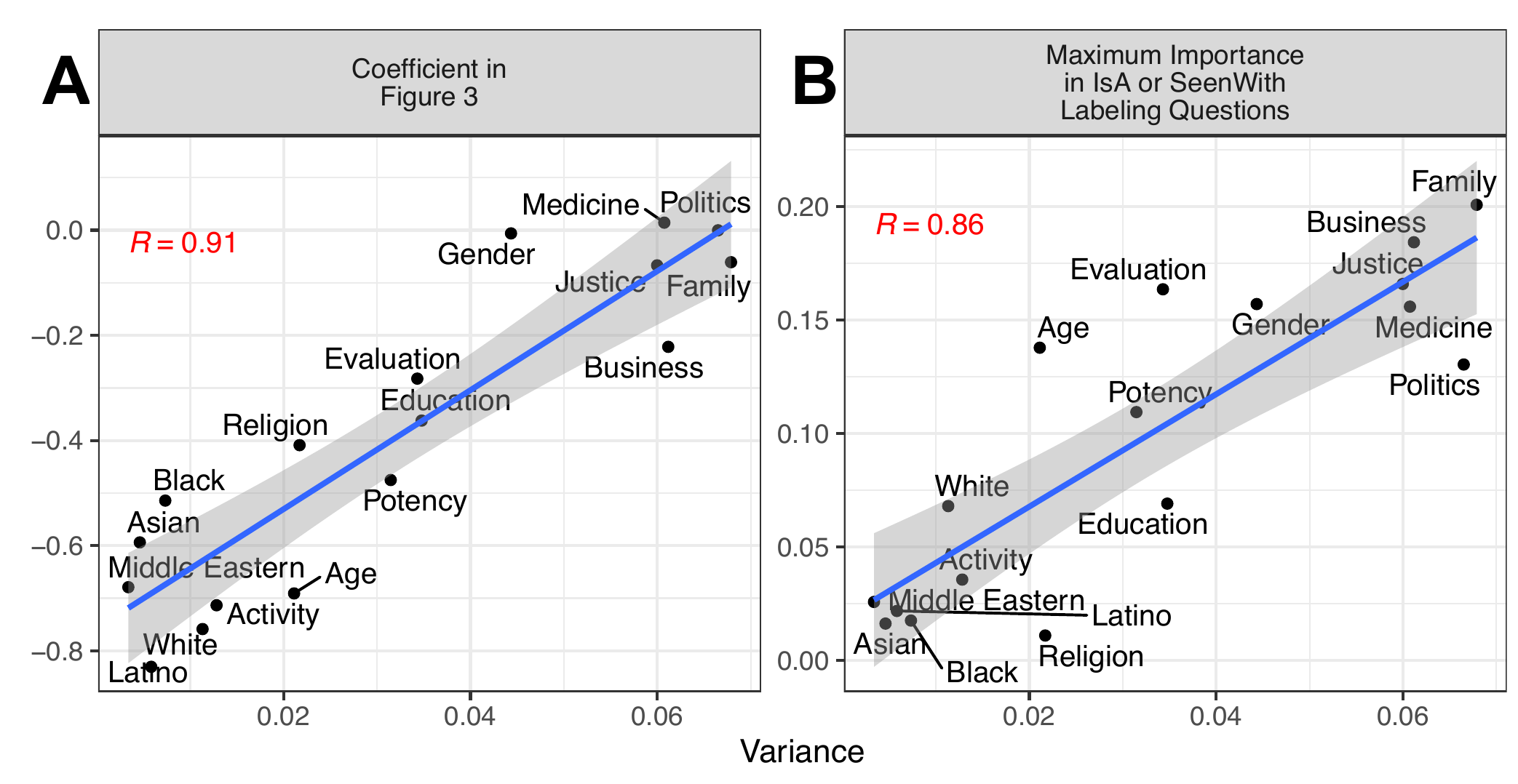}
	\caption{In both A) and B), the y-axis gives the coefficient value for the regression presented in Figure 3. In A), the x-axis represents the variance in survey means along the dimension. In B), an estimate of the dimension's importance for identity labeling}
	\label{fig:why_dim}
\end{figure}

Figure~\ref{fig:why_dim}A) shows, as first identified by \citet{kozlowski_geometry_2018}, that much of this is due to the variance of the survey data along that dimension; the correlation between variance and the coefficients in Figure~\ref{fig:rq1} is 0.91.  However, as discussed above, \citet{kozlowski_geometry_2018} study more general concepts on more general dimensions, and note that they have no easy way to connect their observations to any critical social processes. In contrast, here, Figure~\ref{fig:why_dim}B) shows a significant positive correlation between variance in the survey data along a dimension (and hence measurement accuracy) and that dimensions' importance in explaining patterns of labeling in our identity labeling task. Embedding-based measures of beliefs about identities, we therefore show, are most likely to reflect traditional survey measures particularly when those beliefs are salient for identity labeling.   

\begin{figure}
	\centering
	\includegraphics[width=.48\textwidth]{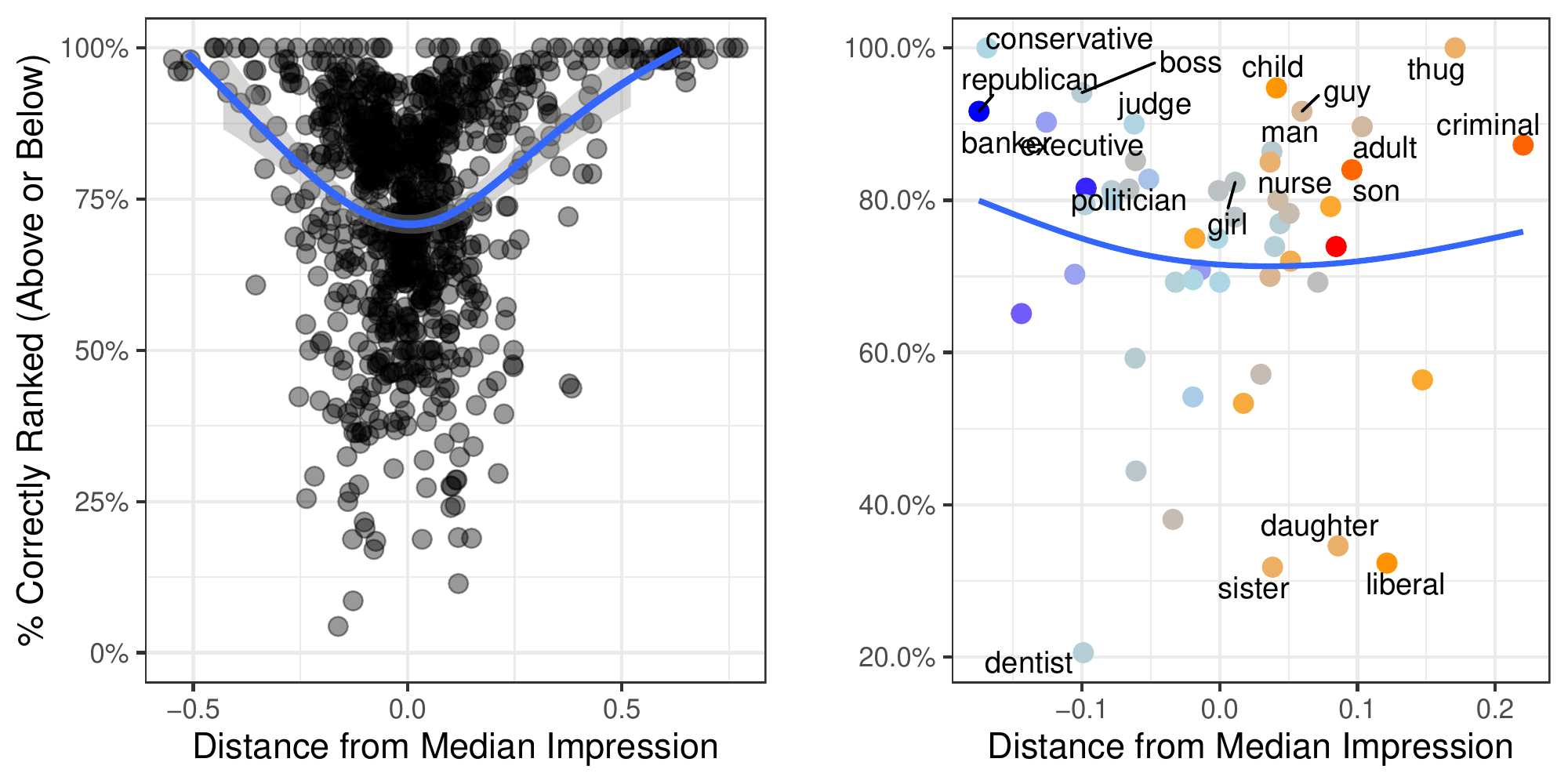}
	\caption{Results for our belief-level outcome (y-axis) as a belief's distance from the dimension-level median (x-axis) varies for all dimensions (left) and only the ``Black'' dimension (right). Points on the right are colored by variance, with higher (lower) variance estimates colored blue (red) }
	\label{fig:impression}
\end{figure}

Critically, then, results for biases in word embeddings are tied not only to the salience of dimensions in general social life, but also to the identities selected for measurement.  Selecting only heavily racialized and non-gendered identities, for example, might well reverse the positions of racialized dimensions and gender in Figure~\ref{fig:why_dim}.  This makes it all the more critical to identify theoretically-driven concepts--- salience in labeling, and variance in measurement--- that move beyond measures of specific identities on specific dimensions to help us understand what is measurable and what is not, particularly when survey data is not available.

\subsection{Belief-level results}

As with the dimension-level results, we find that embedding-based measures are generally accurate predictors of survey-based measures for specific beliefs. On average, 74.9\% of the beliefs   collected for this paper are correctly ranked, as are 82.1\%, 72.0\%, and 71.4\% of the beliefs from \citet{bolukbasi_man_2016-2}, \citet{smith-lovin_interpreting_2015}, and \citet{agarwal-etal-2019-word}, respectively. One caveat to keep in mind, however, is that we focus only on the single best embedding measurement approach for each source/dimension combination. 

Regardless, as with the dimension-level results, there is considerable variance at the belief level. Some of this variance (approximately 32\%, see the Appendix for full regression results ) can be explained by the factors we consider. The strongest explanation we find to explain ranking accuracy, reflected in the left-hand plot in Figure~\ref{fig:impression}, is the distance of the survey-based belief measure from the median on its dimension. At the extremes, ranking accuracy is almost perfect. Because extreme observations are also most likely to be low variance---for example, consider that beliefs at the most extreme values of a scale \emph{must} have zero variance---a more general claim can be made: word embedding-based measures accurately capture our most extreme and agree-upon beliefs about people, but show significant unexplained (at least by us) variance for more neutral and/or less-agreed upon beliefs.

This variance is on display in the right-hand plot in Figure~\ref{fig:impression}, which gives results for the blackness dimension. The embedding-based measure captures with perfect accuracy racialized perceptions of the identities thug and criminal, but not, e.g., liberal, which is similar along the other explanatory factors we consider here. As far as we are aware, it remains an open question as to why this is the case.

\section{Conclusion}

In this paper, we asked, can we trust measures of beliefs about people derived from word embeddings? We find the answer to be yes, at least on average. Depending on one's perspective, this could be good or bad. From a cultural studies/social psychological perspective, this positive correlation further validates efforts to use  word embeddings to study perceptions of people historically, at scale, and in context. On the other hand, from the ``bias'' perspective, this suggests that a vast array of social biases are encoded in embeddings.

However, we also find that some beliefs---specifically, extreme beliefs on salient dimensions--- are easier to measure than others. More generally, across four datasets, we find that \emph{what} we measure is more important than \emph{how} we  measure it. Again, two different perspectives on this are needed. With respect to the study of culture and human stereotypes, we may be safest in studying only the most extreme results from embedding models, as has been done by, e.g., \citet{spirling2019word}. 

From the bias perspective, given the rash of recent work on debiasing word embeddings, our results suggest that much more attention needs to be paid to how we are evaluating these approaches.  Currently, upstream evaluations of debiasing are centered almost exclusively on occupational identities on gender, where some of the most salient social biases we know of exist \cite{ridgeway_framed_2011}. Others have argued that removing these salient beliefs may not remove gender information from embeddings \cite{gonen_lipstick_2019}.  But  \citeauthor{gonen_lipstick_2019}'s \citeyear{gonen_lipstick_2019} argument relies on a technical deficiency of existing approaches. We can make a similar critique by simply changing what is being measured.  For example, the correlation between gender beliefs and the gender direction in the Hard-Debiased embeddings of \citet{bolukbasi_man_2016-2} is 0.05 (p = .84) using identities in their data, \emph{and 0.4 (p \textless .05) using the identities in our data.}  

Similarly, removing gender bias does not remove bias on other dimensions.  For example, while \citet{sweeney_transparent_2019} show that the NumberBatch embeddings harbor the least gender bias, we find that they are the only embedding to show consistently high correlations with age, leading to the potential for ageism downstream. More generally, stereotypes  exist along a \emph{network} of beliefs \cite{freeman_dynamic_2011} reflecting unwarranted \emph{correlations} between many dimensions \cite{ridgeway_framed_2011}; we must therefore be careful not to expect that removing meaning along one dimension will expel social biases from our models.


\section{Acknowledgements}

K.J. was supported by NSF IIS-1939579. This research was supported in part by a Seed Grant from SUNY. We thank Lynn Smith-Lovin, Lisa Friedland, and Yuhao Du for comments on earlier versions of this work.

\bibliography{acl2020}
\bibliographystyle{acl_natbib}

\appendix

\section{Embedding Models}

We use twelve publicly available embedding models.  We use all public GloVe \cite{pennington2014glove} models\footnote{\url{https://nlp.stanford.edu/projects/glove/}}, FastText \cite{mikolov2018advances}  models\footnote{\url{https://fasttext.cc/docs/en/english-vectors.html}}, the original Word2Vec \cite{mikolov_efficient_2013} model\footnote{\url{https://code.google.com/archive/p/word2vec/}}, and v19.08 of the NumberBatch \cite{speer2017conceptnet} model.\footnote{\url{https://github.com/commonsense/conceptnet-numberbatch}}

\section{Word Position Measurement Models}
\begin{table*}
	\centering
	\begin{tabular}{p{2.2cm}lp{3.cm}ll}
	{\bf Measure} & {\bf Normalized?} & {\bf Position Measure} & {\bf Direction-Specification} & {\bf Multiclass} \\ \hline
	 \citet{ethayarajh-etal-2019-understanding} & N & $\frac{\langle w, b \rangle}{||b||}$ & Same as \citet{bolukbasi_man_2016-2} & N \\  \hline
	 \citet{kozlowski_geometry_2018} & Y & $\frac{\langle w, b \rangle}{||b|| ||w||}$ & $\sum_{p_i \in P} \frac{p_{i,l} - p_{i,r}}{||P||}$ & N\\  \hline
	 \citet{bolukbasi_man_2016-2} & Y & $\frac{\langle w, b \rangle}{||b|| ||w||}$ & $SVD \left( \text{c} \left( p_{i,j} - \mu_{p_{ij}} \quad  p_i \in P \right) \right)$ & N \\  \hline
	 \citet{swinger_what_2019} & Y & $\text{avg}_{p_{i} \in P} \frac{\langle w, p_{i,l}\rangle}{||w|| ||p_{i,l}||} - \text{avg}_{p_{i} \in P} \frac{\langle w, p_{i,r}\rangle}{||w|| ||p_{i,r}||}$ & N/A & Y \\  \hline
	 \citet{garg_word_2018} & Y & $||w - b_r|| - ||w- b_l||$ & $b_l := \sum_{p_{i} \in p_r} \frac{p_{i}}{||P||} $  & Y \\  \hline
	\end{tabular}
	\caption{Details of the prior work on word position measurement models from which we draw. We use each model listed here, as well as using the approach of \citet{ethayarajh-etal-2019-understanding} but using direction specification as described by \citet{garg_word_2018} and \citet{kozlowski_geometry_2018}. }
	\label{tab:measure}
\end{table*}

Table~\ref{tab:measure} outlines the word position measurement models used in the present work.  The table provides information on the authors of the measure, whether or not embeddings are normalized before analysis, how  words are measured once a direction has been specified, how a direction is specified, and whether or not the method is ``multi-class,'' described further below.  

Notationally, we have tried to remain as close to the original works as possible. Therefore, $w$ is the identity to  be measured, and $b$ is the vector indicating the direction along which it is to be measured. For \citet{garg_word_2018}, $b_l$ and $b_r$ represent words in the left-hand dimension-inducing word set  (e.g. ``man'' and ``him''  for gender) and $b_r$ the right-hand of the dimension-inducing word  sets  (e.g. ``woman'' and ``her''  for gender). The variables $p_{i,l}$  and  $p_{i,r}$ have similar  meanings for \citet{swinger_what_2019} and \citet{bolukbasi_man_2016-2}.

We use the approaches of \citet{garg_word_2018}, \citet{kozlowski_geometry_2018}, \citet{ethayarajh-etal-2019-understanding}, and \citet{swinger_what_2019} exactly as described in the original texts, except for one modification.  In the case where a paired set of terms is required---all cases except \citet{garg_word_2018} and \citet{swinger_what_2019}---and we have a multi-class measurement (e.g. we measure four different dimensions of racialized beliefs), we first identify a \emph{default} dimension and then compare all other dimensions to it. For race, we follow theory on perceptions of default race categories and assign White to be the default race \cite{maclin2001racial}, and the comparison point for White, following \citet{kozlowski_geometry_2018}, to be Black. For the associative dimensions, we select family for the default, and compare family to politics.

In addition, we consider the possibility that the computationally appealing approach from \cite{ethayarajh-etal-2019-understanding} may be improved by using a different direction specification approach. Therefore, we consider two additional word position measurement models, \citet{ethayarajh-etal-2019-understanding} + \citet{garg_word_2018}, and \citet{ethayarajh-etal-2019-understanding} + \citet{kozlowski_geometry_2018}, that are the same as the original model but using the direction specification method in these two papers instead of the method from \citet{bolukbasi_man_2016-2}, as was done in the original paper.

\section{Further Details - Identity Selection for Our Survey Data}

In selecting identities for each cluster, we also ensured that the words selected were a) expressed most frequently as identities, b) were in a standard set of lexical resources and thus common English terms and c) were used relatively frequently. To ensure the identities were used first and foremost as an identity (and not, e.g., as a verb or place name), we first used both the NLP python library \texttt{spacy} and Wordnet to identify any identities for which the dominant sense was a verb (e.g. suspect, accused) and removed these from consideration. To ensure that the identities were in common lexical databases, we removed words which were not in Wordnet as a noun or an adjective. Finally, to ensure that identity words were used frequently, we checked that they were used frequently in \emph{either} a fairly informal medium, Twitter, or in a fairly formal medium, Wikipedia. To check the former case, we use the frequency counts of words from 56M tweets given by \citet{owoputi_improved_2013} and retain only those identities used in more than 2500 tweets. To check Wikipedia, we first extract all 532,051 ``clean'' pages from a Wikipedia dump from December, 2015. A clean page is a page that is not labeled as a stub, that was still active one month after the dump was created, and that also has more than 50 views over 2 year span, where we pull one random hour for each day.

 \section{Further Details - Belief Measurement Data}
 
 \subsection{Measurement}
The slider bar for the affective dimensions gives labels at different points, ranging from ``Infinitely'' to ``Slightly'' on both ends, with a ``Neutral'' option in the middle. The age slider had the following qualitative labels, spaced equally across the slider bar: ``Baby, Child, Teenager, 20s, 30s, 40s, 50s, 60s, 70s, 80s, 90s, \textgreater=100''.  The gender slider had the following labels, spaced equally across the slider: ``Always Male, Mostly Male, Equally Male or Female, Mostly Female, Always Female''. 

For the race/ethnicity beliefs, order of the sliders was randomized, and the starting value for each was set to 20\%. With respect to discussions about the 2020 census, most importantly, demographers have pushed to include Hispanic or Latino as a racial category rather than to split it out into its own separate question.
        
For the associative belief question, presentation order was randomized. The form of the question is drawn from other studies seeking to elicit cognitive associations between a term and a set of other concepts, e.g. from \citet{hill_simlex-999_2014}.  The specific institutions were originally drawn from the clustering used to determine our identities.  However, we added the education and religion institutions after determining they would be necessary for a more complete meaning space, as suggested by the institutional settings with which identities are commonly associated as discussed by \citet{heise_self_2010}.

Finally, pilot testing suggested that respondents became confused when provided with certain identities that had meanings that were used to construct the question - for example, on the race question, respondents became confused when being asked ``Of all white people, what percentage of them do you think are ... [White]?'' We, therefore, removed the gender question from the identities guy, boy, girl, lady, man, and woman, and removed the race question from the identities Asian, White person, Black person, Arab, and Hispanic.

\subsection{Participant Sample}
In total, 252 Mechanical Turk workers completed the survey. These had greater than a 95\% completion rate and had completed over 1,000 HITs. These workers also were located within the United States. In order to ensure this was the case, we leveraged a tool provided by \citet{winter_simplified_2019} to ensure that participants' IP addresses were located within the United States and that they were not using a VPN. We further ensured competency and attention by including two attention check questions. Five respondents were rejected because they failed attention checks. Sample demographics were not collected for Task 1, because we make the deliberate assumption that measurements from surveys of any individuals within a national culture can serve as the foundation for a meaning space.  However, as noted, we do ensure cultural expertise by ensuring participants are native English speakers located in the U.S.

\subsection{Ethical Approval}
       This study was approved by the Institutional Review board of the University at Buffalo.
  
\subsection{Summary Results}\label{app_sec:study2_task1_fullres}
        
        \begin{figure*}[t]
            \centering
            \includegraphics[width=\linewidth]{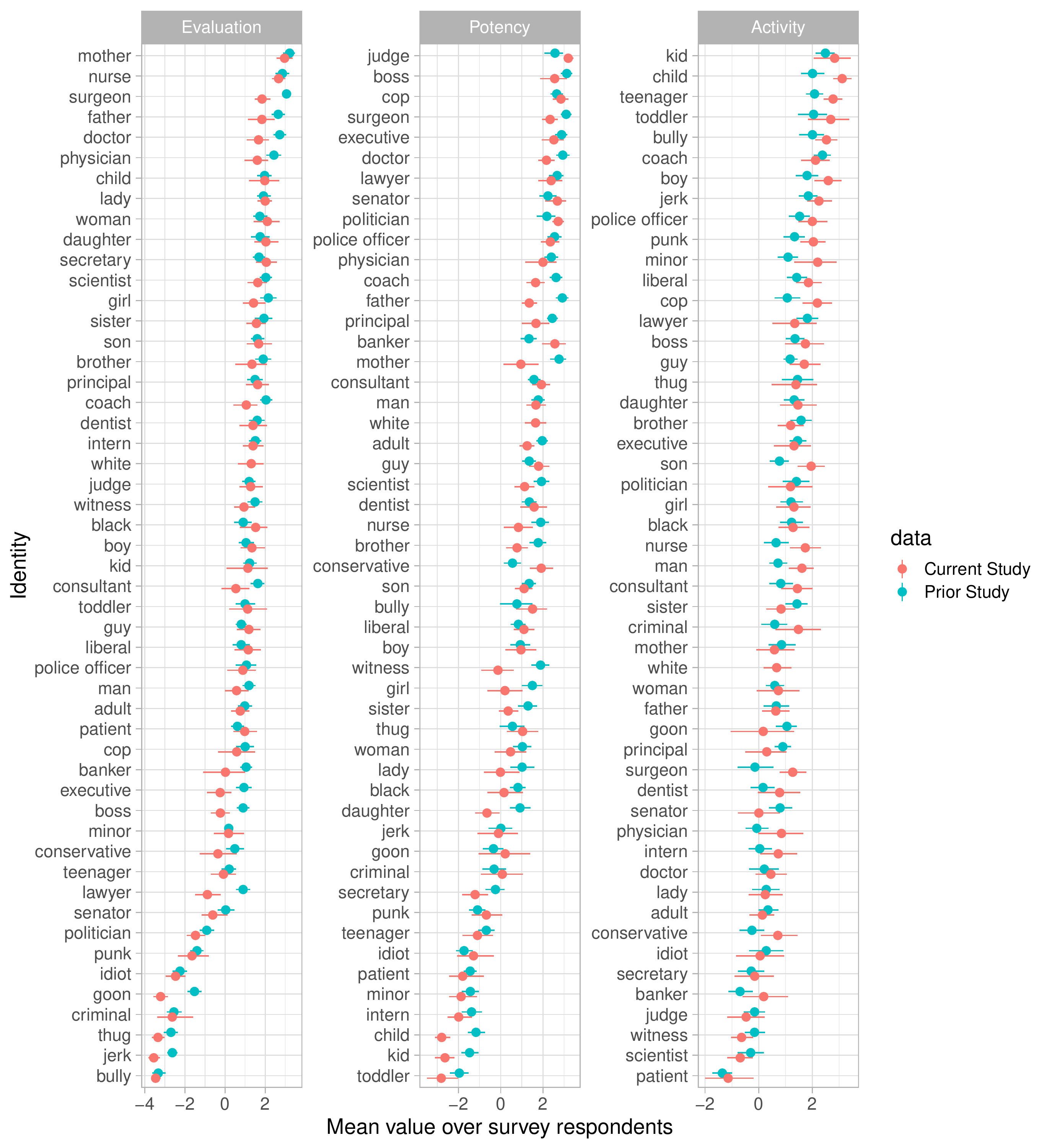}
            \caption{95\% bootstrapped confidence intervals for estimates of the Evaluation, Potency, and Activity dimensions.  Results from the current study are in red, results from a prior study from \citet{smith-lovin_interpreting_2015} are given for comparison in cyan.}
            \label{fig:epa_study2_task2}
        \end{figure*}

Figure~\ref{fig:epa_study2_task2} provides 95\% bootstrapped confidence intervals for the Evaluation, Potency, and Activity dimensions for each identity in the survey.  The measurements in our survey are compared to results from \citet{smith-lovin_interpreting_2015}. The vast majority of our estimates overlap closely with theirs, signifying the broad generality of the measurement tools used by ACT across individuals within a national culture.  Where differences arise, we do not believe one dataset or the other appears to be universally more accurate, and further, given the number of comparisons (171, 3 per each of 57 identities), we should expect even by chance some larger differences in the measurements.
  
    \begin{figure*}[t]
        \centering
        \includegraphics[width=\textwidth]{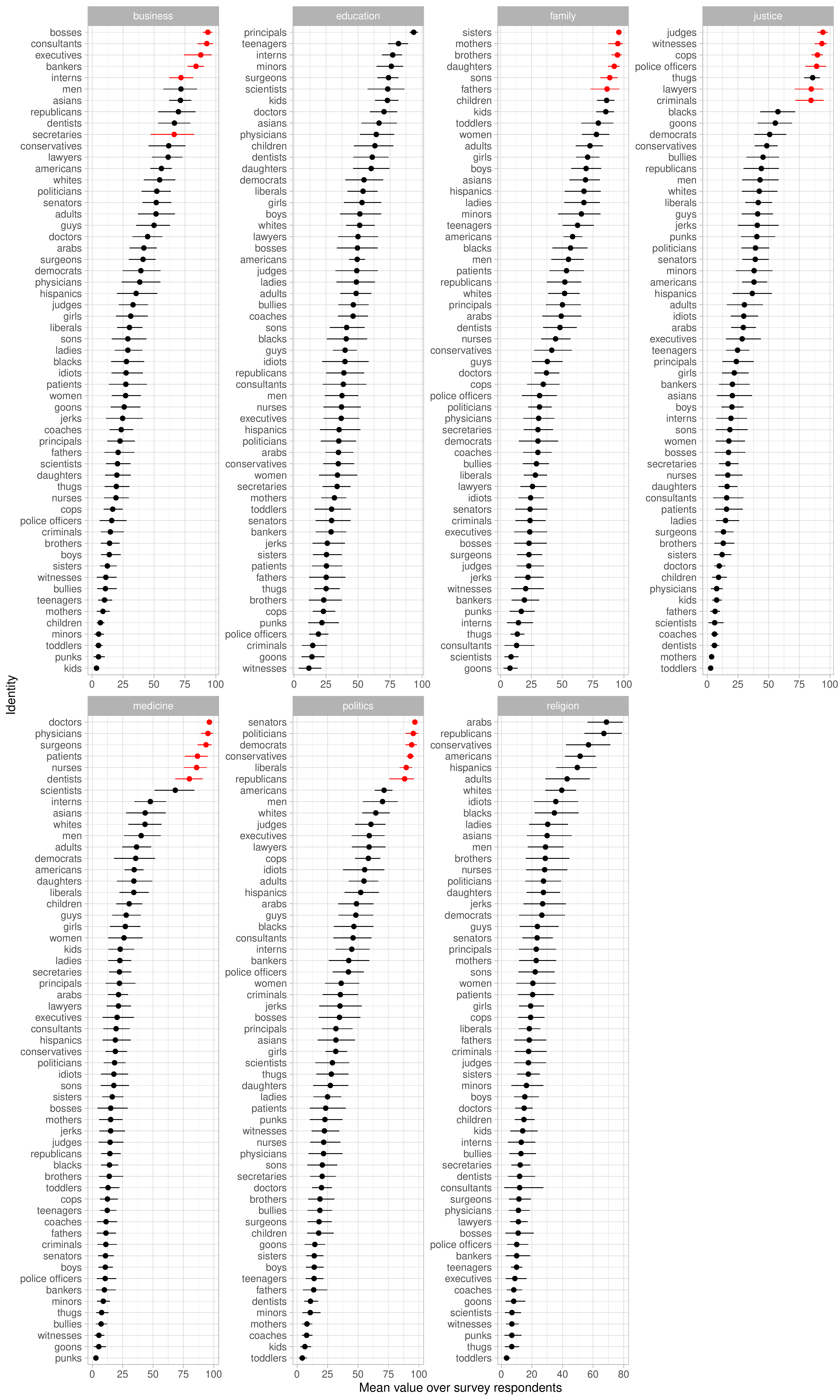}
        \caption{Associations for each identity to each of the institutional settings we study.  Results are given with 95\% bootstrapped confidence intervals, and each institutional setting is a different subplot.  Red estimates represent identities that were assumed to be clustered on the given institutional setting}
        \label{fig:association_study2_task2}
    \end{figure*}
    
Figure~\ref{fig:association_study2_task2} provides full results for associative meanings.  The figure shows that in general, identities assumed to cluster within a particular institution were rated as having the highest associations with that institution. However, it is also clear that other identities were also strongly aligned with the various institutions in ways that did not follow exactly from the construction of our identity set. For example, the identity thug was included as a negative affect term, but was perceived to have a strong association to the judicial institution.
        
        \begin{figure*}[t]
            \centering
            \includegraphics[width=\linewidth]{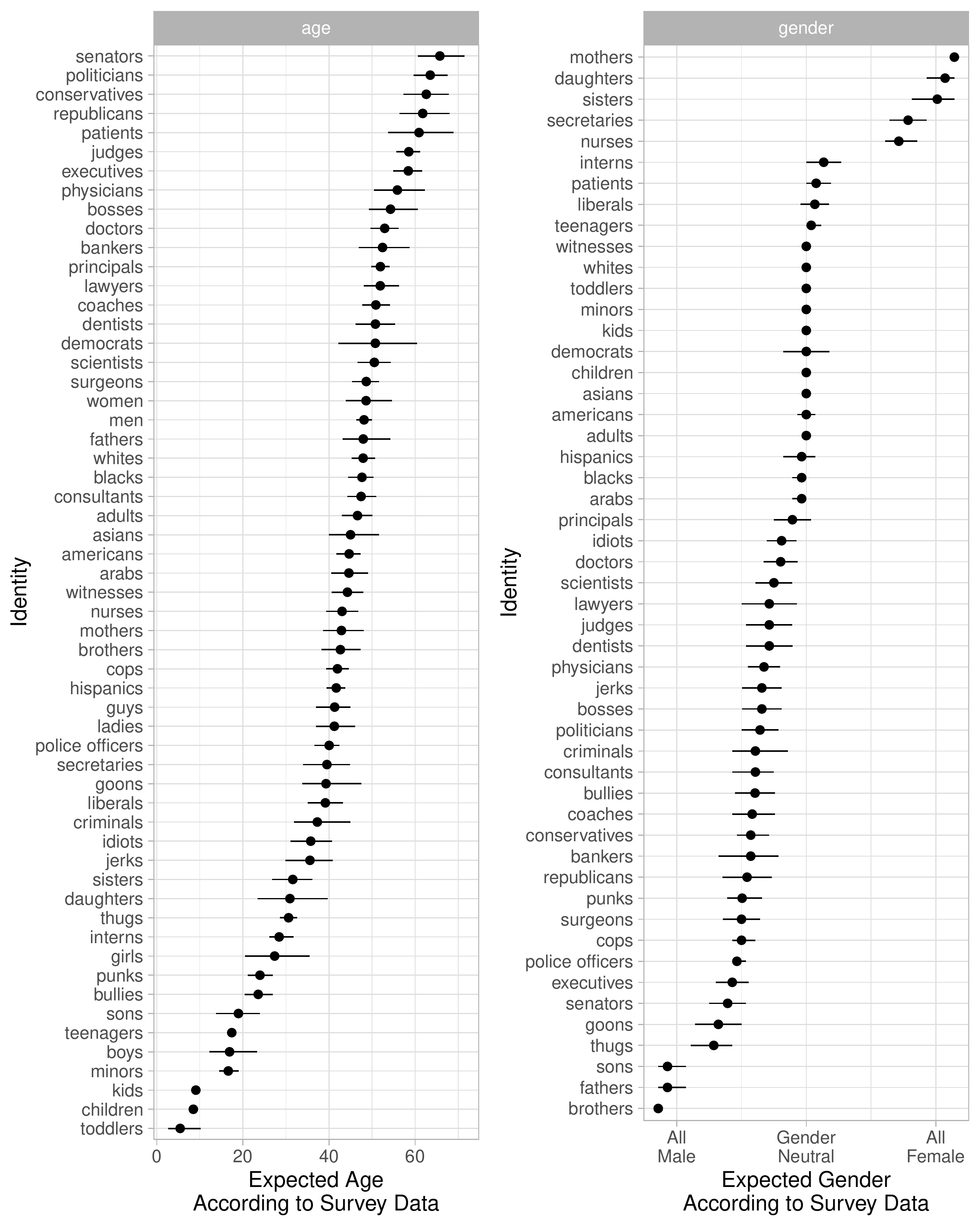}
            \caption{On the left, 95\% confidence intervals for the expected age of each identity.  On the right, 95\% confidence intervals for the expected gender of each identity, assuming a continuous dimension for gender. Note the gender labels are slightly shifted inward so labels are readable, meaning some points extend beyond the label itself.}
            \label{fig:age_gender_study2_task2}
        \end{figure*}

Figure~\ref{fig:age_gender_study2_task2} provides full results for the expected age and gender of each identity. Note that some identities with a denotative meaning aligned with these dimensions were included, because pilot tests did not suggest confusion for these identities. We therefore attempted to include as many identities as possible in the actual measurement.

        \begin{figure*}[t]
            \centering
            \includegraphics[width=\linewidth]{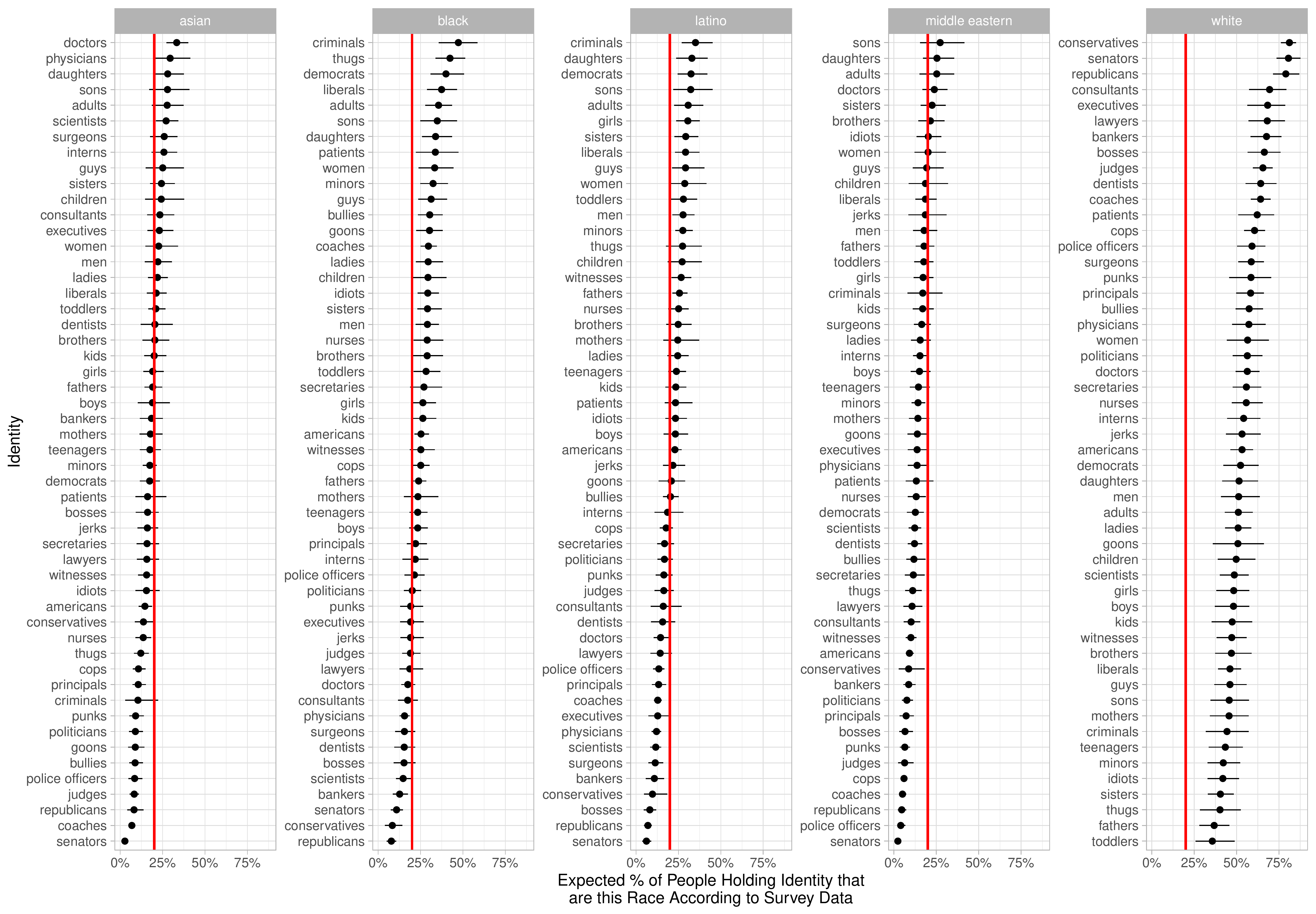}
            \caption{95\% confidence intervales for the perceived percentage of each identity that are also of a given race. The five racial/ethnic categories we consider are represented as separate subplots, the red line at 20\% gives the expected value of all races were equally represented for the identity.}
            \label{fig:race_study2_task2}
        \end{figure*}

Figure~\ref{fig:race_study2_task2} provides full results for the race question we asked.  All identities were assumed by participants to be more White than expected by chance.  Recall that the racial question was not posed for the denotatively aligned racial identities we studied.

\section{Further Details - Identity Labeling Task }\label{app_sec:study2_task2_description}

        \subsection{Participant Sample}
       
We collected valid data from 402 Mechanical Turk workers were located in the United States, had greater than a 95\% completion rate and had completed over 1,000 HITs. To assess accuracy for respondents, we randomly sampled 5 questions from each respondent and ensured that answers did not appear to be entered randomly.  The sample's gender was 53.7\% female, and 45.6\% male (.7\% did not say). A total of 89.4\% had at least some college or vocational training, 53.8\% of the sample had completed at least a Bachelor's degree, and 16.1\% had a post-graduate degree. Almost all (96.7\%) of the sample were born in and had lived between 75-100\% of their life in the United States.  With respect to age, 9.0\% of the sample was aged 18-24, 17.2\% aged 25-29, 33.3\% aged 30-39, and 40.0\% aged 40 or older (.5\% did not say). Finally, the sample was largely white, 83.3\% of participants were White or Caucasian.  

\subsection{Ethical Approval}
        The survey carried out was approved by the Institutional Review board of Carnegie Mellon University.

\subsection{Additional Implementation Details}

As noted in the text, and replicated here for clarity, from the 57 identities in Table~\ref{tab:id_s2}, we create survey questions for the identity labeling task as follows: for a given identity, we generate 14 random sets of the 56 other identities; each set contains four identities. We then generate one IsA and one SeenWith question for each of these sets, where these four identities constitute the possible answers to the question, and the given identity is used in the question text. This process is then repeated ten times for each identity. This process generates exactly ten questions for each of the 3,192\footnote{57*56=3,192} identity pairs for each type of question. 
        
The intention was, therefore, to have exactly ten questions for each identity pair for each question type where the first identity in the pair is shown in the question and the second identity in the pair is shown as a possible answer. In each case, by construction, the other possible answers were randomly selected. Unfortunately, our survey suffered from a bug with the Qualtrics software used, where the option to present questions an even number of times fails in unclear cases.  Due to this error, some of our identity pairs were not seen exactly ten times. Specifically, 3.4\% were asked less than 6 times, 40\% were asked less than ten times, and 40.4\% were asked more than ten times.   While this does not affect our analyses, because they do not rely on any exact number of questions per identity pairing, it is important to note for purposes of any future work with the dataset.

Such issues aside, the process described generated 15,960 questions.  These questions produced a total of 16,080 responses (a small number of questions---120, or 0.7\%---were asked more than once in attempts to address the Qualtrics bug) that were split evenly, 40 questions per respondent.
    
\subsection{Results}

\begin{figure}[ht]
	\centering
	\includegraphics[width=.5\textwidth]{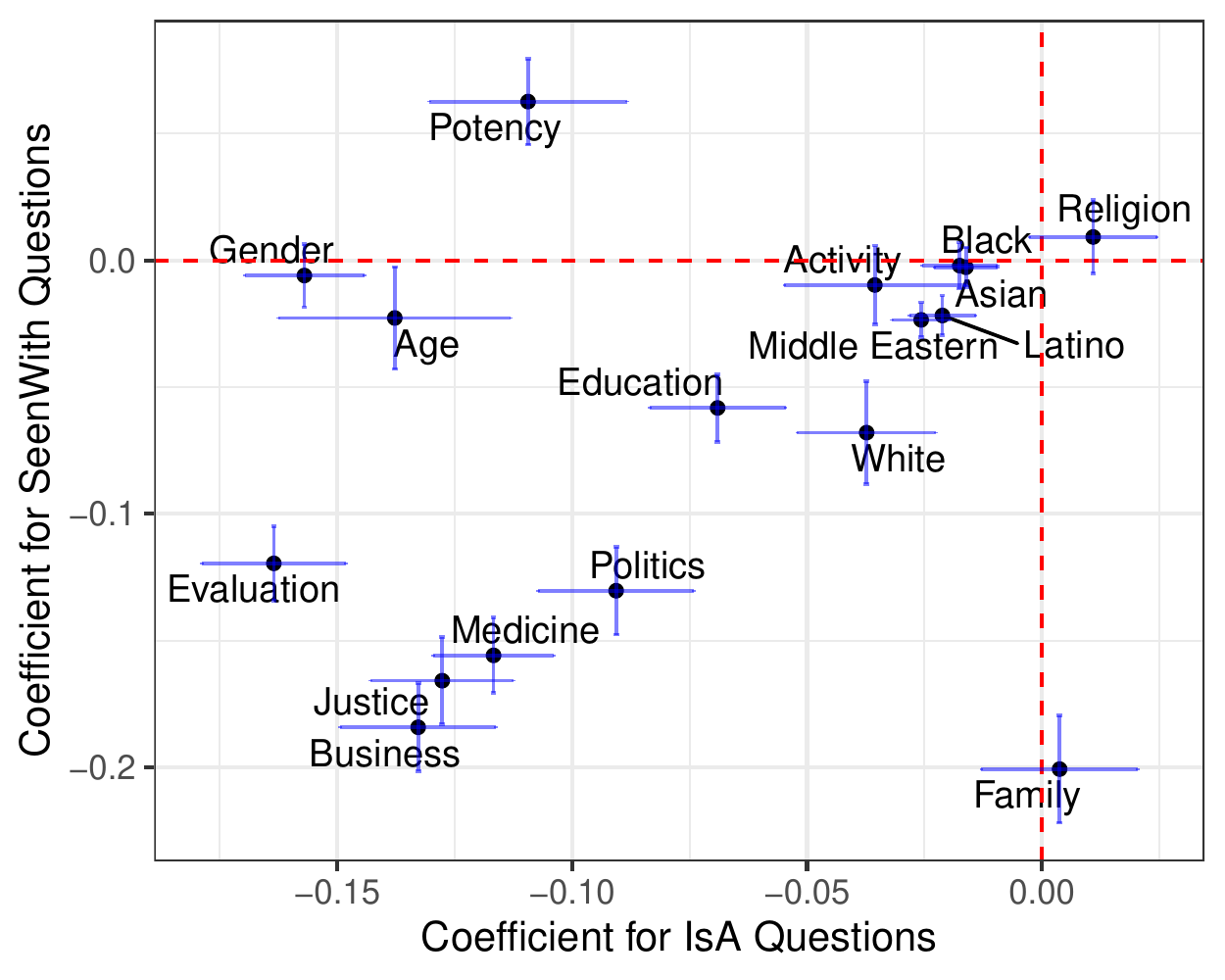}
	\caption{Coefficients for the regression models described in the main text for IsA and SeenWith questions for each dimension. Blue bars represent error bars for each dimension on IsA questions (horizontal) and SeenWith (vertical). Red dotted lines represent zero, any error bars that cross this line are not significant}
	\label{fig:salience}
\end{figure}

Figure~\ref{fig:salience} presents full results for the regression models described in the text to identify the importance of each dimension for identity labeling.
    
\section{Additional Details - RQ2}

\begin{figure*}[ht]
	\centering
	\includegraphics[width=\linewidth]{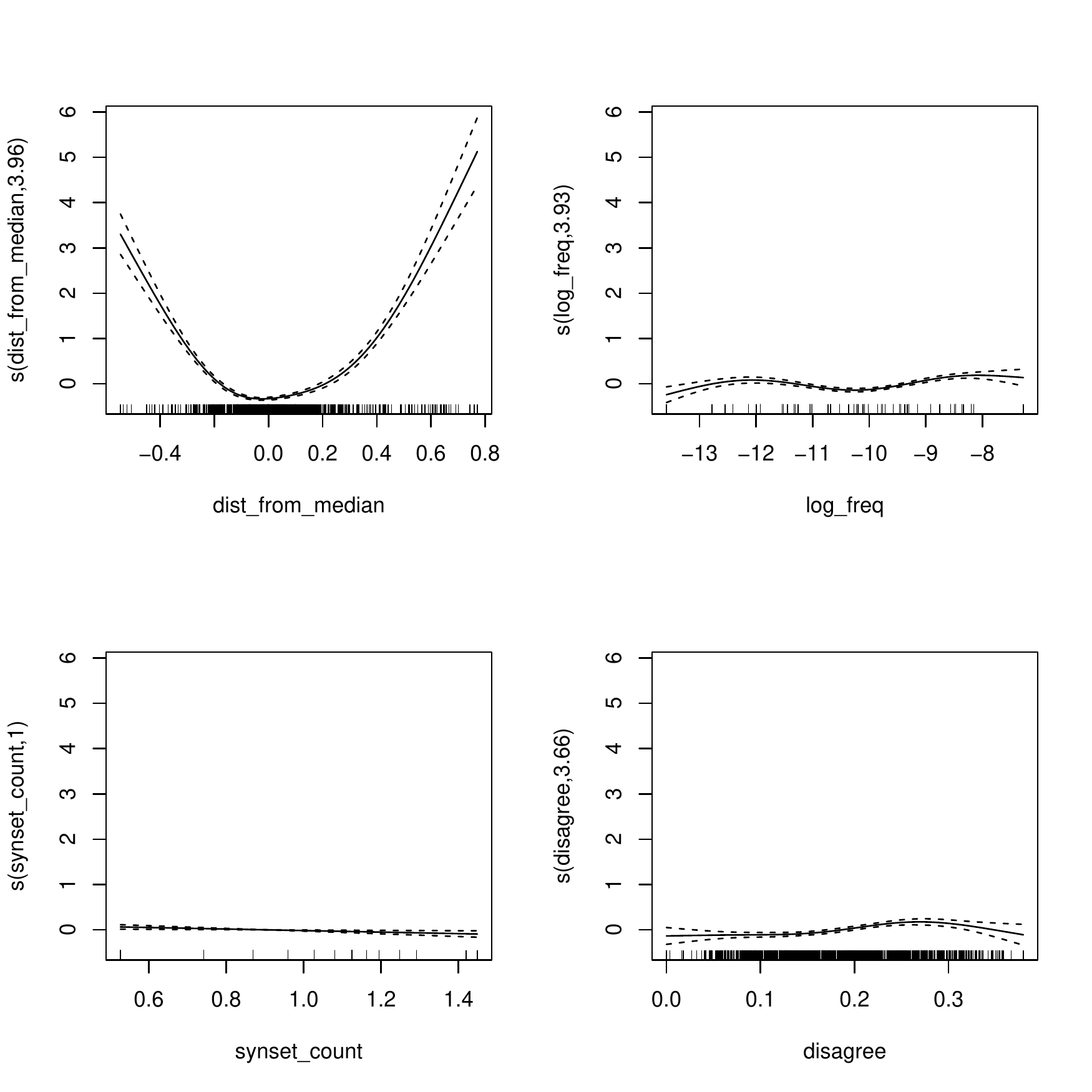}
	\caption{Regression results for survey data from this paper on the belief-level measure}
	\label{fig:salience2}
\end{figure*}

\begin{figure*}[ht]
	\centering
	\includegraphics[width=\linewidth]{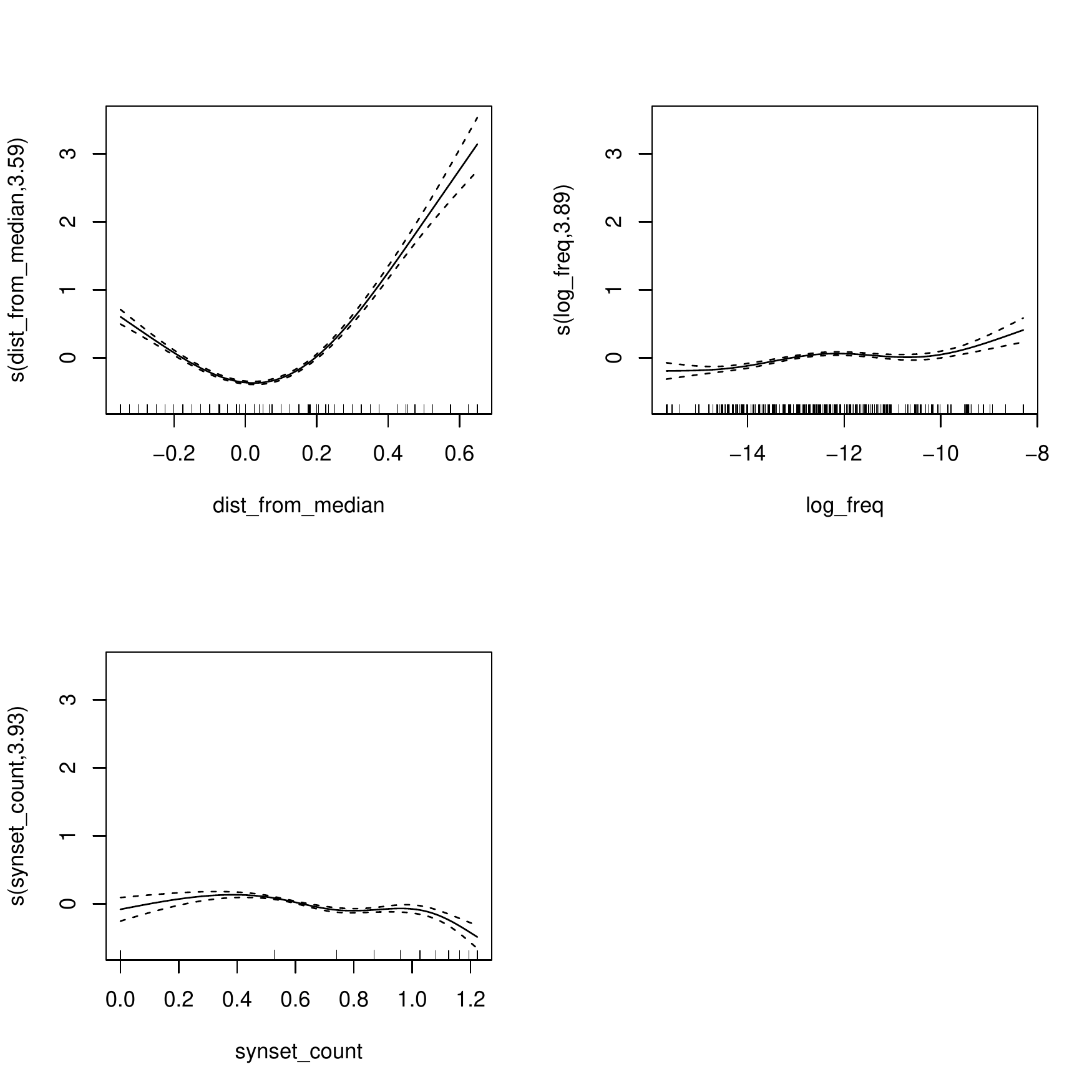}
	\caption{Regression results for survey data from \citet{bolukbasi_man_2016-2} on the belief-level measure}
	\label{fig:salience3}
\end{figure*}

\begin{figure*}[ht]
	\centering
	\includegraphics[width=\linewidth]{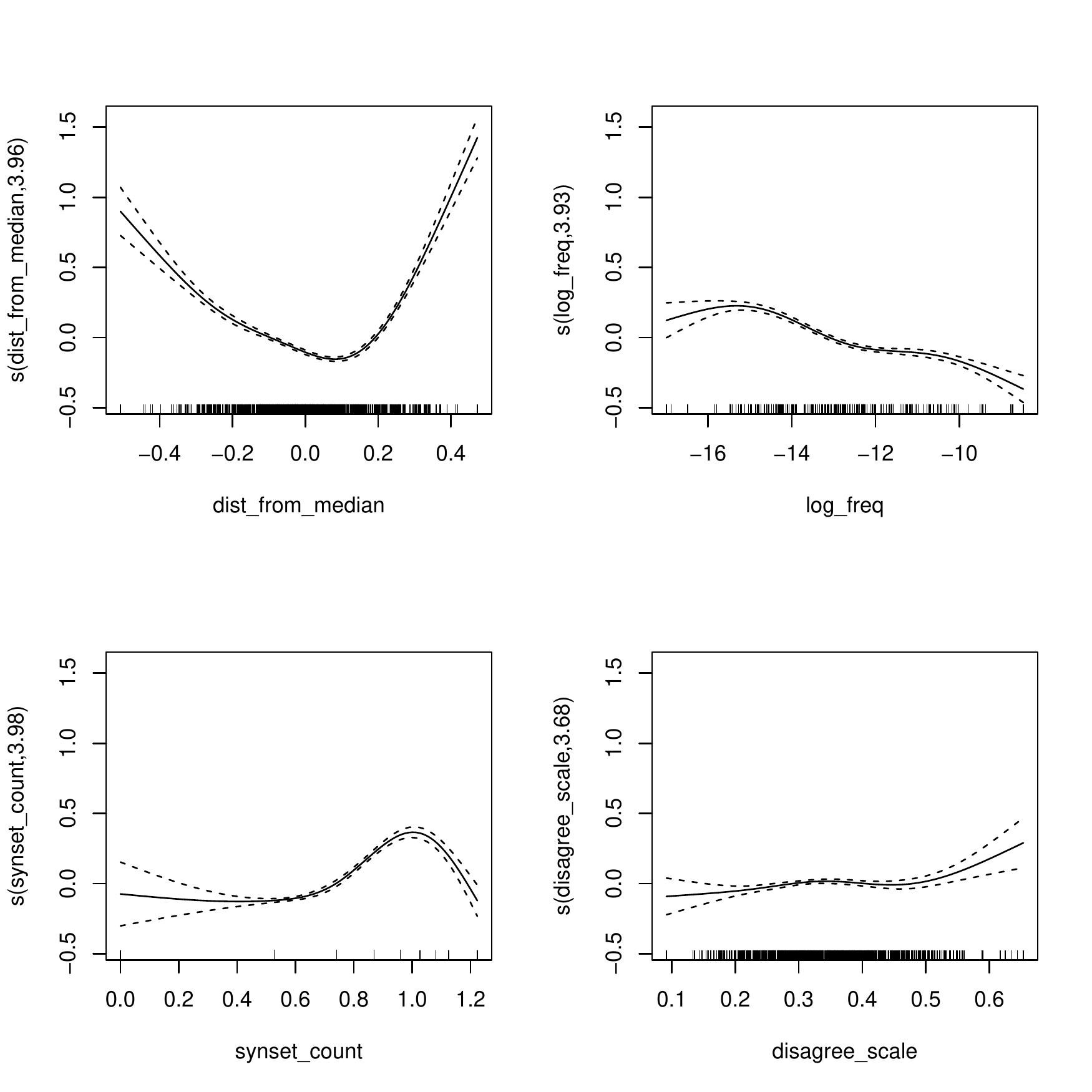}
	\caption{Regression results for survey data from \citet{agarwal-etal-2019-word}  on the belief-level measure}
	\label{fig:salience4}
\end{figure*}

\begin{figure*}[ht]
	\centering
	\includegraphics[width=\linewidth]{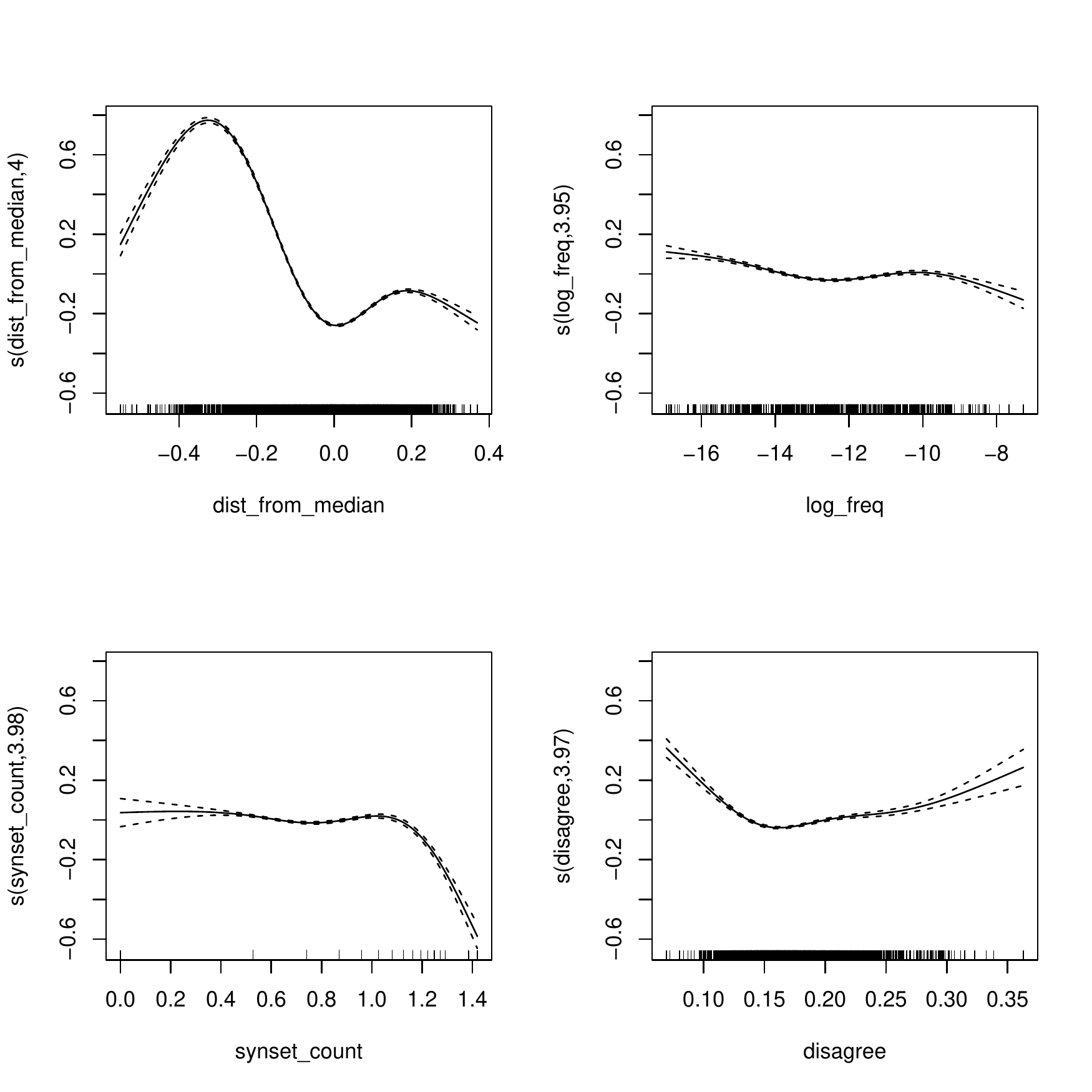}
	\caption{Regression results for survey data  from \citet{smith-lovin_interpreting_2015} on the belief-level measure}
	\label{fig:salience5}
\end{figure*}

Figures~\ref{fig:salience2}-\ref{fig:salience5} present results from generalized additive models with the four dependent variables described in the main text on the rank-level outcome variable at the belief level. The models explain 31.8\%, 34.2\%, 16.9\%, and 21.6\% of the deviance for the data from this paper, \citet{bolukbasi_man_2016-2}, \citet{agarwal-etal-2019-word}, and \citet{smith-lovin_interpreting_2015}, respectively.  Across the four datasets, the only consistent predictor is the distance of the survey-based belief measure from the median. Note, however, that in the data from \citet{smith-lovin_interpreting_2015}, this pattern does not hold at the extremes. Further analyses suggests this is due to a small number of outliers on the extremes of the Evaluation dimension, and does not appear to reflect any interesting trend worth additional consideration.  Additionally, we note that, the authors of \citet{bolukbasi_man_2016-2} could only provide us with a mean per-identity estimate, and thus no information on the variance of those estimates is used in our results.

\end{document}